\documentclass[journal]{IEEEtran}
\usepackage{graphicx}
\usepackage{epsfig}
\usepackage{latexsym}
\usepackage{amsfonts}
\usepackage{here}
\usepackage{rawfonts}
\usepackage[latin1]{inputenc}
\usepackage[T1]{fontenc}
\usepackage{calc}
\usepackage{url}
\usepackage{enumerate}
\usepackage{color}
\usepackage[tbtags]{amsmath}
\usepackage{amssymb}
\usepackage{upref}
\usepackage{epic,eepic}
\usepackage{times}
\usepackage{dsfont}
\usepackage{comment}
\usepackage{cite}

\usepackage{graphicx}
\usepackage[font={small}]{caption}
\usepackage[font={small}]{subcaption}

\usepackage{stfloats}
\usepackage{mathtools}
\usepackage{amsmath}

\usepackage{algorithm}
\usepackage{algpseudocode}

\newtheorem{proposition}{Proposition}

\ifCLASSINFOpdf
\else
\fi
\begin{document}

\title{Hierarchical Federated ADMM}

\author{{Seyed Mohammad Azimi-Abarghouyi, Nicola Bastianello,~\IEEEmembership{Member,~IEEE}, Karl H. Johansson,~\IEEEmembership{Fellow,~IEEE}, and Viktoria Fodor,~\IEEEmembership{Member,~IEEE}}

	\thanks{The authors are with the School of Electrical Engineering and Computer Science and Digital Futures, KTH Royal Institute of Technology, Stockholm, Sweden (Emails: $\bigl\{$seyaa, nicolba, kallej, vjfodor$\bigr\}$@kth.se). 
		
	The work of N.B. and K.H.J. was partially supported by the European Union's Horizon Research and Innovation Actions programme under grant agreement No. 101070162, and partially by Swedish Research Council Distinguished Professor Grant 2017-01078 Knut and Alice Wallenberg Foundation Wallenberg Scholar Grant.}
}

\maketitle

\vspace{-15pt}
\begin{abstract}
In this paper, we depart from the widely-used gradient descent-based hierarchical federated learning (FL) algorithms to develop a novel hierarchical FL framework based on the alternating direction method of multipliers (ADMM). Within this framework, we propose two novel FL algorithms, which both use ADMM in the top layer: one that employs ADMM in the lower layer and another that uses the conventional gradient descent-based approach. The proposed framework enhances privacy, and experiments demonstrate the superiority of the proposed algorithms compared to the conventional algorithms in terms of learning convergence and accuracy. Additionally, gradient descent on the lower layer performs well even if the number of local steps is very limited, while ADMM on both layers lead to better performance otherwise.

\end{abstract}
\vspace{0pt}
\vspace{-5pt}
\section{Introduction}
Federated learning (FL) is gaining popularity for tasks involving data that is either sensitive or expensive to collect, making privacy a key concern \cite{mcmahan}. Additionally, FL accelerates the learning process by allowing parallel computations across multiple clients. Traditionally, FL has been implemented using a central server \cite{mcmahan}, but recent developments propose using a hierarchy of servers for improved scalability \cite{cast}. In this hierarchical FL model, two layers of data aggregation take place. The lower-layer "edge" aggregation is performed at edge servers, each serving a distinct set of clients, during intra-set iterations. In contrast, the top-layer "cloud" aggregation is executed at a cloud server during inter-set iterations.

The main motivation for utilizing hierarchical architectures often arises from the wide geographic distribution of clients. In wireless networks, this structure can improve transmission quality over wireless channels \cite{wen} and increase the efficiency of the learning process by localizing certain operations, which conserves communication resources and reduces time \cite{gupta}. Additionally, clustering clients can help manage heterogeneity in device and network capabilities \cite{lin, wu}, limit data traffic to specific administrative regions or social groups \cite{zhou}, and align with the topology of mobile networks or computing infrastructures \cite{zhang, gupta}.

Hierarchical FL employing quantization methods has been proposed in \cite{letaief, ourotherjournal}, with its integration with over-the-air computation investigated in \cite{ourconf, ourjournal}. Another bandwidth-limited approach involves using pruning to reduce the scale of the neural network \cite{prun1}. Additionally, network optimization problems such as client selection, resource allocation, and clustering are discussed in \cite{bennis, tony, energy_resource, feduc}. All of this previous research builds on conventional FL theory based on gradient descent optimization, where model or gradient parameters are transmitted and aggregated, typically by averaging, at both the top and lower layers. This conventional approach reveals the entire knowledge of the model by directly transmitting its parameters. 
\begin{figure}[tb!]
	\vspace{0pt}
	\centering
	\includegraphics[width =3.6in]{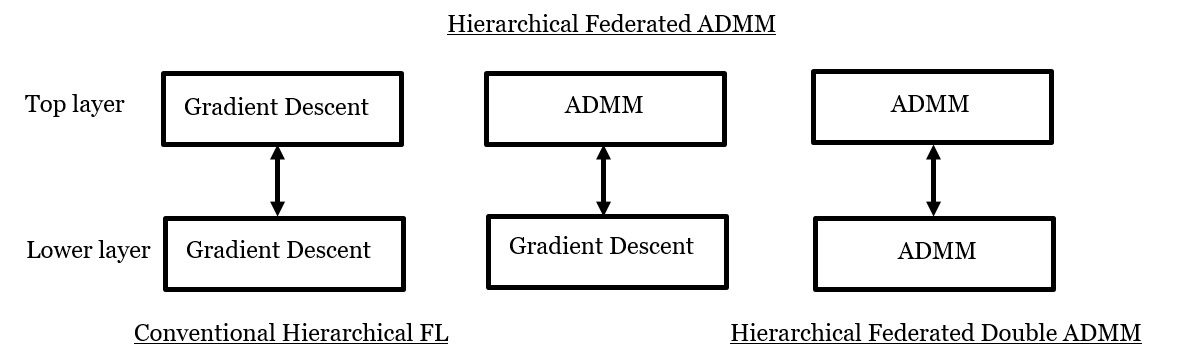} 
	\vspace{-7pt}
	\caption{Modular schematic of FL algorithms}
	\vspace{-15pt}
\end{figure}

Recently, an FL approach based on the alternating direction method of multipliers (ADMM) \cite{boyd} within a single client set has been proposed \cite{fedadmm, shashi}. This approach not only enhances privacy by obscuring model parameters during transmission but also significantly improves learning performance in terms of convergence and accuracy. Hierarchical structures could also benefit from such advanced distributed optimization methods to spare communication or local computing resources, or to improve accuracy. 

In this paper, we propose a systematic approach to build hierarchical structures, that can be reused for different optimization methods. We demonstrate it for the case of ADMM. That is, we consider ADMM for the top layer of the hierarchy, integrating all clients across various sets into a unified training problem, and derive two possible optimization approaches for the lower layer: hierarchical federated double ADMM ({\fontfamily{lmtt}\selectfont {{HierF2ADMM}}}), which uses ADMM at the lower layer, and hierarchical federated ADMM ({\fontfamily{lmtt}\selectfont {{HierFADMM}}}), which uses gradient descent and averages model parameters at the lower layer, similar to conventional FL. A schematic representation of the conventional and proposed hierarchical FL algorithms is depicted in Fig. 1. This illustration demonstrates how our approach allows for the modular integration of two optimization methods within FL, a concept introduced in this work. Our experimental results demonstrate the superiority of these ADMM-based hierarchical FL algorithms over conventional hierarchical FL in various scenarios, including both i.i.d. and non-i.i.d. data distributions. Additionally, while our {\fontfamily{lmtt}\selectfont {{HierF2ADMM}}} generally outperforms our {\fontfamily{lmtt}\selectfont {{HierFADMM}}} method, the latter excels when the number of local training steps is limited, and consequently, an appropriate method can be selected according to the resources of the clients.
\vspace{-5pt}
\section{Network Model and Learning Problem
}
Assume there are one cloud server and $C$ edge servers with distinct client sets $\{\mathcal{V}^c\}_{c=1}^{C}$. Each set $\mathcal{V}^c$ includes $N_c$ clients, with each client $k$ possessing a local dataset $\mathcal{D}_{kc}$. Then, the global objective function is
\begin{align}
\label{lossfunction}
F(\mathbf{w}) = \sum_{c=1}^{C}\frac{D_c}{D} F_c(\mathbf{w}),
\end{align}
where $D = \sum_{c=1}^{C} D_c$ with each $D_c = \sum_{k=1}^{N_c} D_{kc}$, and $D_{kc} = |{\cal D}_{kc}|, \forall k, c$. In \eqref{lossfunction}, the intra-set objective function corresponding to set $c$ is
\begin{align}
F_c(\mathbf{w}) = \sum_{k=1}^{N_c} \frac{D_{kc}}{D_c} f_{kc}(\mathbf{w}),
\end{align}
where the local objective function at client $k \in {\cal V}^c$ is
\begin{align}
\label{localloss}
f_{kc}(\mathbf{w}) =  \frac{1}{D_{kc}}\sum_{\xi\in {\cal D}_{kc}}^{}\ell(\mathbf{w},\xi),
\end{align}
where $\ell(\mathbf{w},\xi)$ is the sample-wise objective function that measures the
prediction error of $\mathbf{w}$ on a sample $\xi$.
Therefore, the main problem of the learning process is to find the optimal model parameter vector $\mathbf{w}^*$, which minimizes the global objective function as
\begin{align}
\label{opt_FL}
\mathbf{w}^* = \min_{\mathbf{w}} F(\mathbf{w}).
\end{align}
\section{Conventional Hierarchical Federated Learning
}
Conventional hierarchical FL, referred to as {\fontfamily{lmtt}\selectfont {HierFed}} \cite{letaief}, builds on gradient descent as its optimization foundation and applies the following two-step process at each global iteration $t+1$ to solve \eqref{opt_FL}.
 
\subsubsection{Step 1} The edge servers transmit their intra-set model parameters, $\mathbf{w}_c^t, \forall c$, to the cloud server, where the following \textit{cloud aggregation} occurs.

\begin{align}
\mathbf{w}^{t+1} = \frac{1}{D} \sum_{c=1}^{C} {D_c}\mathbf{w}_c^t.
\end{align}
\subsubsection{Step 2} Each client updates its local model parameters as $\mathbf{w}_{kc}^{0,0, t+1} = \mathbf{w}^{t+1}, \forall c, k$. Next, $\tau$ intra-set iterations are performed at each set $c$ as follows. During each intra-set iteration $i$, each client $k \in {\cal V}^c$ updates its local model parameters (local training) using an $L$-step gradient descent as follows, $l = \left\{0,\ldots,L-1\right\}$:
\begin{align}
\label{local}
&\mathbf{w}_{kc}^{l+1,i,t+1} = \mathbf{w}_{kc}^{l,i,t+1} - \mu \nabla f_{kc}(\mathbf{w}_{kc}^{l,i,t+1}),
\end{align}
where $\mu$ is the learning rate. Upon completion, all clients $\forall k \in {\cal V}^c$ send their local model parameters to the edge server $c$, where the following \textit{edge aggregation} occurs.
\begin{align}
\mathbf{w}_c^{i+1,t+1} = \frac{1}{D_c} \sum_{k=1}^{N_c}D_{kc}\mathbf{w}_{kc}^{L,i,t+1}.
\end{align}
Then, each client updates its parameters as $\mathbf{w}_{kc}^{0,i, t+1} = \mathbf{w}^{i,t+1}_c, \forall k \in {\cal V}^c$ for the next intra-set iteration. After $\tau$ intra-set iterations, the intra-set model parameters are updated as $\mathbf{w}_c^{t+1} = \mathbf{w}_c^{\tau,t+1}$.

\section{Hierarchical Federated ADMM}
We can reformulate the main problem \eqref{opt_FL} at the top layer as
\begin{align}
\label{re_opt}
\min_{\mathbf{w}, \mathbf{W}} \sum_{c=1}^{C}\frac{D_c}{D} F_c(\mathbf{w}_c), \ \text{subject to} \ \mathbf{w}_c = \mathbf{w}, \forall c,
\end{align}
where $\mathbf{W} = \left\{\mathbf{w}_1, \ldots, \mathbf{w}_C\right\}$ represents the collection of all intra-set model parameter vectors from all sets. Based on the general ADMM approach as in \cite{boyd}, we propose {\fontfamily{lmtt}\selectfont {{HierFADMM}}} with the following three steps in every iteration $t+1$ to solve \eqref{re_opt}, which shares a similar structure with \cite[Equation~(13)]{fedadmm}.
\begin{align}
\label{my_admm}
\left\{
\begin{aligned}
&\mathbf{w}^{t+1} = \arg \min_{\mathbf{w}} {\cal L}\left(\mathbf{w}, \mathbf{W}^t, \boldsymbol \Pi^t \right), \\
&\mathbf{w}^{t+1}_c = \arg \min_{\mathbf{w}_c} {\cal L}_c\left(\mathbf{w}^{t+1}, \mathbf{w}_c, \boldsymbol \pi_c^t \right), \forall c,\\
&{\boldsymbol \pi}^{t+1}_c = {\boldsymbol \pi}^{t}_c + \sigma_c \left(\mathbf{w}^{t+1}_c - \mathbf{w}^{t+1}\right), \forall c,
\end{aligned}
\right.
\end{align}
where ${\cal L}$ represents the global augmented Lagrangian function of the problem \eqref{re_opt} and ${\cal L}_c$ denotes its corresponding intra-set function for each set $c$, as defined later in \eqref{global_log} and \eqref{step_2}, respectively. Additionally, $\boldsymbol{\pi}_c$ is the Lagrangian multiplier and $\sigma_c > 0$ is a constant associated with the intra-set function of set $c$. We also introduce $\boldsymbol \Pi = \left\{\boldsymbol \pi_1, \dots, \boldsymbol \pi_C\right\}$. In the following, {\fontfamily{lmtt}\selectfont {{HierFADMM}}} employs a federated learning approach to address each step in \eqref{my_admm}.
\subsubsection{Step 1}
According to \eqref{re_opt}, the global augmented Lagrangian function is 
\begin{align}
\label{global_log}
{\cal L}\left(\mathbf{w}, \mathbf{W}, \boldsymbol \Pi \right) = \sum_{c=1}^C {\cal L}_c\left(\mathbf{w}, \mathbf{w}_c, \boldsymbol \pi_c \right),
\end{align}
where the intra-set function of each set $c$ is given by
\begin{align}
\label{step_2}
&{\cal L}_c\left(\mathbf{w}, \mathbf{w}_c, \boldsymbol \pi_c \right) = \\ &= \frac{D_c}{D}F_c(\mathbf{w}_c) + \langle \mathbf{w}_c - \mathbf{w}, \boldsymbol{\pi}_c \rangle + \frac{\sigma_c}{2} \|\mathbf{w}_c - \mathbf{w}\|^2  \nonumber\\
&= \frac{D_c}{D} \sum_{k=1}^{N_c} \frac{D_{kc}}{D_c}f_{kc}(\mathbf{w}_c) + \langle \mathbf{w}_c - \mathbf{w}, \boldsymbol{\pi}_c \rangle + \frac{\sigma_c}{2} \|\mathbf{w}_c - \mathbf{w}\|^2 \nonumber\\
&= \frac{D_c}{D} \sum_{k=1}^{N_c}\biggl( \frac{D_{kc}}{D_c}f_{kc}(\mathbf{w}_c) + \frac{D}{D_c N_c}\langle \mathbf{w}_c - \mathbf{w}, \boldsymbol{\pi}_c \rangle + \nonumber\\
& \quad  +\frac{\sigma_c}{2}\frac{D}{D_c N_c} \|\mathbf{w}_c - \mathbf{w}\|^2\biggr) = \frac{D_c}{D}\sum_{k=1}^{N_c} \frac{D_{kc}}{D_c}\tilde f_{kc}(\mathbf{w}_c,\mathbf{w}, \boldsymbol{\pi}_c), \nonumber
\end{align}
where
\begin{align}
\tilde f_{kc}(\mathbf{w}_c,\mathbf{w}, \boldsymbol{\pi}_c) &= f_{kc}(\mathbf{w}_c) + \frac{D}{D_{kc}N_c}\langle \mathbf{w}_c - \mathbf{w}, \boldsymbol{\pi}_c \rangle \nonumber\\&+ \frac{\sigma_c}{2}\frac{D}{D_{kc}N_c} \|\mathbf{w}_c - \mathbf{w}\|^2.
\end{align}
Taking derivative from ${\cal L}\left(\mathbf{w}, \mathbf{W}^t, \boldsymbol \Pi^t \right)$ with respect to $\mathbf{w}$ and equating the resulting expression to zero, similar to \cite{fedadmm}, we obtain 
\begin{align}
\label{my_cloud}
\mathbf{w}^{t+1} = \frac{1}{\sum_{c=1}^{C} \sigma_c} \sum_{c=1}^{C} \left(\sigma_c \mathbf{w}^{t}_c + \boldsymbol \pi_c^t\right).
\end{align}
This is \textit{cloud aggregation} in {\fontfamily{lmtt}\selectfont {{HierFADMM}}}. For this, each edge server $c$ transmits $\sigma_c \mathbf{w}^{t}_c + \boldsymbol \pi_c^t$ to the cloud server. Since an intentionally concealed version of the intra-set model parameters $\mathbf{w}^{t}_c$ is shared with the cloud server, our approach enhances privacy compared to the conventional approach, which requires transmitting the intra-set model parameters directly.
\subsubsection{Step 2} From \eqref{step_2}, the second optimization problem in \eqref{my_admm} at the lower layer is 
\begin{align}
\label{step2}
&\mathbf{w}^{t+1}_c = \arg \min_{\mathbf{w}_c} \sum_{k=1}^{N_c} \frac{D_{kc}}{D_c}\tilde f_{kc}(\mathbf{w}_{kc},\mathbf{w}^{t+1}, \boldsymbol{\pi}_c^{t}), \nonumber\\& \text{subject to}\ \mathbf{w}_{kc} = \mathbf{w}_c, \forall k \in {\cal V}^c.
\end{align}

Given that each edge server $c$ lacks access to local objective functions $f_{kc}$ in $\tilde f_{kc}$ for all $k \in {\cal V}^c$, we propose, in this section, the following federated learning approach based on gradient descent to execute the distributed minimization in \eqref{step2} with an iterative manner. In the next section, we will introduce an alternative approach based on ADMM. We remark that, beyond the privacy preservation of $f_{kc}$, approximating the solution of \eqref{step2} is also necessary as this update does not have a closed form solution in general.

Each client updates its local model parameters as $\mathbf{w}_{kc}^{0,0, t+1} = \mathbf{w}^{t+1}, \forall c, k$. Following this, $\tau$ intra-set iterations are conducted at each set $c$. In each of these iterations $i$, every client $k \in {\cal V}^c$ updates its local model parameters via an $L$-step gradient descent process as $l = \left\{0,\ldots,L-1\right\}$:
\begin{align}
\label{my_local}
\mathbf{w}_{kc}^{l+1,i,t+1} &= \mathbf{w}_{kc}^{l,i,t+1} - \mu \nabla\tilde f_{kc}(\mathbf{w}_{kc}^{l,i,t+1} ,\mathbf{w}^{t+1},\boldsymbol{\pi}_c^t)\nonumber\\&= \mathbf{w}_{kc}^{l,i,t+1} -\mu \biggl( \nabla f_{kc}(\mathbf{w}_{kc}^{l,i,t+1}) + \frac{D}{D_{kc}N_c} \boldsymbol{\pi}_c^t  +\nonumber\\& \qquad  + {\sigma_c}\frac{D}{D_{kc}N_c} (\mathbf{w}_{kc}^{l,i,t+1} - \mathbf{w}^{t+1})\biggr),
\end{align}
where the gradient descent direction is determined by taking the derivative of the local objective function $\tilde{f}_{kc}$ as shown in \eqref{step2}. Please note that setting $\boldsymbol \pi_c^t = \mathbf{0}$ and $\sigma_c = 0$ in \eqref{my_local} yields the same local update \eqref{local} as the conventional gradient descent. After completing the $L$ steps, all clients $\forall k \in {\cal V}^c$ transmit their local model parameters to the edge server $c$, where \textit{edge aggregation} then takes place.
\begin{align}
\label{my_edge}
\mathbf{w}^{i+1,t+1}_c = \frac{1}{D_c} \sum_{k=1}^{N_c}D_{kc}\mathbf{w}_{kc}^{L,i,t+1}.
\end{align}
Subsequently, for the upcoming intra-set iteration $i+1$, each client $k \in {\cal V}^c$ sets its parameters as $\mathbf{w}_{kc}^{0,i, t+1} = \mathbf{w}^{i,t+1}_c$. Once $\tau$ intra-set iterations are completed, the intra-set model parameters are updated as $\mathbf{w}^{t+1}_c = \mathbf{w}_c^{\tau,t+1}, \forall c$.

\subsubsection{Step 3} Finally, the last step in \eqref{my_admm} involves updating the Lagrangian multipliers, which can be performed locally at each client $k \in {\cal V}^c$ and the edge server $c$ as
\begin{align}
\label{my_multiple}
{\boldsymbol \pi}^{t+1}_c = {\boldsymbol \pi}^{t}_c + \sigma_c \left(\mathbf{w}^{t+1}_c - \mathbf{w}^{t+1}\right).
\end{align}

\begin{algorithm}
	\small
	\caption{{\fontfamily{lmtt}\selectfont {HierFADMM}} algorithm}
	\begin{algorithmic}
		\vspace{-1pt}
		\State Select hyperparameters $\mu$ and $\sigma_c,\forall c$. 
		\State Initialize $\mathbf{w}^0$ and $\boldsymbol \pi_c^0,\forall c$.\
		\vspace{0pt}
		\State \textbf{for} global iteration $t=1,...,T$ \textbf{do}
		\vspace{0pt}
		\State \hspace{10pt}Each client updates its model by $\mathbf{w}^{t}$. 
		\vspace{0pt}
		\State \hspace{10pt}\textbf{for} intra-set iteration $i=1,...,\tau$ \textbf{do}
		\vspace{0pt}
		\State \hspace{20pt}Each client $k \in \mathcal{V}^c$ updates its local model $\mathbf{w}_{kc}^{l,i,t}$ for $l = \left\{0,\ldots,L-1\right\}$ according to \eqref{my_local}.
		\vspace{0pt}
		\State \hspace{20pt}Each edge server $c$ obtains its intra-set model $\mathbf{w}^{i+1,t}_c$ according to \eqref{my_edge}.
		\vspace{0pt}
		\State \hspace{20pt}Each client $k \in \mathcal{V}^c$ updates as $\mathbf{w}_{kc}^{0,i+1, t} = \mathbf{w}^{i+1,t}_c$.
		\vspace{0pt}
		\State \hspace{10pt}\textbf{end}
		\State \hspace{10pt}Each edge server $c$ updates as $\mathbf{w}^{t}_c = \mathbf{w}_c^{\tau,t}$.
		\State \hspace{10pt}Each edge server $c$ and its intra-set clients update ${\boldsymbol \pi}^{t}_c$ according to \eqref{my_multiple}.

		\State \hspace{10pt}Cloud server obtains global model $\mathbf{w}^{t+1}$ according to \eqref{my_cloud}.
		\State \textbf{end}
	\end{algorithmic}
	
\end{algorithm}

\vspace{-5pt}
\section{Hierarchical Federated Double ADMM}
In this section, we introduce an alternative hierarchical FL algorithm, {\fontfamily{lmtt}\selectfont {HierF2ADMM}}, which maintains the same learning structure as {\fontfamily{lmtt}\selectfont {HierFADMM}}, particularly retaining \textit{Steps 1 and 3}. However, in \textit{Step 2}, we employ a different iterative federated learning approach based on ADMM to solve the distributed minimization \eqref{step2} at the lower layer. During each intra-set iteration $i+1$, this can be characterized as follows.
\begin{align}
\label{my2_admm}
\left\{
\begin{aligned}
&\mathbf{w}^{i+1,t+1}_c = \arg \min_{\mathbf{w}_c^{t+1}} \tilde {\cal L}_c\left(\mathbf{w}_c^{t+1}, \mathbf{W}_c^{i,t+1}, \boldsymbol \Pi_c^{i,t+1} \right), \\
&\mathbf{w}^{i+1,t+1}_{kc} = \arg \min_{\mathbf{w}_{kc}^{t+1}} \tilde {\cal L}_{kc}\left(\mathbf{w}^{i+1,t+1}_c, \mathbf{w}_{kc}^{t+1}, \boldsymbol \pi_{kc}^{i,t+1} \right) \forall k \in {\cal V}^c,\\
&{\boldsymbol \pi}^{i+1,t+1}_{kc} = {\boldsymbol \pi}^{i,t+1}_{kc} + \sigma_{kc} \left(\mathbf{w}^{i+1,t+1}_{kc} - \mathbf{w}^{i+1,t+1}_c\right) \forall k \in {\cal V}^c,
\end{aligned}
\right.
\end{align}
where $\mathbf{W}_c = \left\{\mathbf{w}_{1c}, \ldots, \mathbf{w}_{N_cc}\right\}$ and $\boldsymbol{\Pi}_c = \left\{\boldsymbol{\pi}_{1c}, \ldots, \boldsymbol{\pi}_{N_cc}\right\}$ represent the collections of all model parameter vectors and all Lagrangian multipliers associated with the clients in set $c$. Additionally, $\sigma_{kc} > 0$ is a constant associated with client $k \in {\cal V}^c$. According to \eqref{step2}, the augmented Lagrangian functions in \eqref{my2_admm} are expressed as follows.
\begin{align}
\label{tilde_step1}
\tilde{\cal L}_c\left(\mathbf{w}_c^{t+1}, \mathbf{W}_c^{t+1}, \boldsymbol \Pi_c^{t+1} \right) = \sum_{k=1}^{N_c} \tilde {\cal L}_{kc}\left(\mathbf{w}_{c}^{t+1}, \mathbf{w}_{kc}^{t+1}, \boldsymbol \pi_{kc}^{t+1} \right),
\end{align}
and 
\begin{align}
\label{step_12}
&\tilde {\cal L}_{kc}\left(\mathbf{w}_c^{t+1}, \mathbf{w}_{kc}^{t+1}, \boldsymbol \pi_{kc}^{t+1} \right) = \frac{D_{kc}}{D_c} \tilde f_{kc}(\mathbf{w}_{kc}^{t+1},\mathbf{w}^{t+1}, \boldsymbol{\pi}_c^{t})+ \nonumber\\
& \qquad  + \langle \mathbf{w}_{kc}^{t+1} - \mathbf{w}_c^{t+1}, \boldsymbol{\pi}_{kc}^{t+1} \rangle + \frac{\sigma_{kc}}{2} \|\mathbf{w}_{kc}^{t+1} - \mathbf{w}_c^{t+1}\|^2 \nonumber\\& = \frac{D_{kc}}{D_c} \biggl(\tilde f_{kc}(\mathbf{w}_{kc}^{t+1},\mathbf{w}^{t+1}, \boldsymbol{\pi}_c^{t})+ \frac{D_c}{D_{kc}}\langle \mathbf{w}_{kc}^{t+1} - \mathbf{w}_c^{t+1}, \boldsymbol{\pi}_{kc}^{t+1} \rangle \nonumber\\& \qquad + \frac{\sigma_{kc}}{2} \frac{D_c}{D_{kc}}\|\mathbf{w}_{kc}^{t+1} - \mathbf{w}_c^{t+1}\|^2\biggr).
\end{align}
Following the approach outlined in Section IV, the three steps in \eqref{my2_admm} can be executed as follows. First, all the clients in the set $c$ send a linear combination of their local model parameters and Lagrangian multipliers from the previous iteration $i$ as $\sigma_{kc} \mathbf{w}_{kc}^{L,i,t+1} + \boldsymbol \pi_{kc}^{i,t+1}$ to the edge server $c$ for \textit{edge aggregation} as
\begin{align}
\label{my2_edge}
\mathbf{w}^{i+1,t+1}_c = \frac{1}{\sum_{k=1}^{N_c} \sigma_{kc}} \sum_{k=1}^{N_c} \left(\sigma_{kc} \mathbf{w}_{kc}^{L,i,t+1} + \boldsymbol \pi_{kc}^{i,t+1}\right),
\end{align}
which minimizes \eqref{tilde_step1} as in the first step of \eqref{my2_admm}. Then, each client $k$ sets its parameters to $\mathbf{w}_{kc}^{0,i+1,t+1} = \mathbf{w}^{i+1,t+1}_c$, and subsequently updates them through an $L$-step gradient descent process to minimize \eqref{step_12} in the second step of \eqref{my2_admm} as follows $l = \left\{0,\ldots,L-1\right\}$:
\begin{align}
	\label{my2_local}
	&\mathbf{w}_{kc}^{l+1,i+1,t+1} = \nonumber\\
	&=\mathbf{w}_{kc}^{l,i+1,t+1} - \mu \frac{D_{c}}{D_{kc}} \nabla\tilde {\cal L}_{kc}\left(\mathbf{w}_c^{i+1,t+1}, \mathbf{w}_{kc}^{l,i+1,t+1}, \boldsymbol \pi_{kc}^{i,t+1} \right) \nonumber \\ &= \mathbf{w}_{kc}^{l,i+1,t+1} - \mu\biggl(\nabla\tilde f_{kc}(\mathbf{w}_{kc}^{l,i+1,t+1} ,\mathbf{w}^{t+1},\boldsymbol{\pi}_c^t)+  \frac{D_c}{D_{kc}}\boldsymbol{\pi}_{kc}^{i,t+1}  \nonumber \\ & \qquad + {\sigma_{kc}}\frac{D_c}{D_{kc}} (\mathbf{w}_{kc}^{l,i+1,t+1} - \mathbf{w}_c^{i+1,t+1})\biggr) \nonumber \\ &= \mathbf{w}_{kc}^{l,i+1,t+1} -\mu \biggl( \nabla f_{kc}(\mathbf{w}_{kc}^{l,i+1,t+1}) + \frac{D}{D_{kc}N_c} \boldsymbol{\pi}_c^t \nonumber \\ & \qquad + {\sigma_c}\frac{D}{D_{kc}N_c} \bigl(\mathbf{w}_{kc}^{l,i+1,t+1} - \mathbf{w}^{t+1}\bigr)+  \frac{D_c}{D_{kc}}\boldsymbol{\pi}_{kc}^{i,t+1} \nonumber \\ & \qquad  + {\sigma_{kc}}\frac{D_c}{D_{kc}} (\mathbf{w}_{kc}^{l,i+1,t+1} - \mathbf{w}_c^{i+1,t+1})\biggr).
\end{align}
This is enabled since the client $k$ has access to the function $\tilde {\cal L}_{kc}$. Please note that by setting $\boldsymbol{\pi}_{kc}^{i,t+1} = \mathbf{0}$ and $\sigma_{kc} = 0$ in \eqref{my2_local}, the local update becomes equivalent to that of {\fontfamily{lmtt}\selectfont {HierFADMM}}. Then, as in the third step of \eqref{my2_admm}, each client $k$ updates its local Lagrangian multiplier as
\begin{align}
\label{my2_multiple}
{\boldsymbol \pi}^{i+1,t+1}_{kc} = {\boldsymbol \pi}^{i,t+1}_{kc} + \sigma_{kc} \left(\mathbf{w}^{L,i+1,t+1}_{kc} - \mathbf{w}^{i+1,t+1}_c\right).
\end{align}
After completing $\tau$ intra-set iterations, the intra-set model parameters are updated as $\mathbf{w}^{t+1}_c = \mathbf{w}_c^{\tau,t+1}$ for all $c$.

In {\fontfamily{lmtt}\selectfont {HierF2ADMM}}, both edge servers and all clients conceal their model parameters during transmission, thereby achieving higher privacy compared to {\fontfamily{lmtt}\selectfont {HierFADMM}}.

Finally, it is important to note that the computational complexity of the three algorithms, {\fontfamily{lmtt}\selectfont HierFed}, {\fontfamily{lmtt}\selectfont HierFADMM}, and {\fontfamily{lmtt}\selectfont HierF2ADMM}, is equivalent. This is because the local updates \eqref{local}, \eqref{my_local}, and \eqref{my2_local} all involve the same gradient calculation.

\begin{algorithm}
	\small
	\caption{{\fontfamily{lmtt}\selectfont {HierF2ADMM}} algorithm}
	\begin{algorithmic}
		\vspace{-1pt}
		\State Select hyperparameters $\mu$ and $(\sigma_c, \sigma_{kc}, \forall k \in {\cal V}^c), \forall c$. 
		\State Initialize $\mathbf{w}^0$ and $(\boldsymbol \pi_c^0, \boldsymbol \pi_{kc}^{0,0},\forall k \in {\cal V}^{c}),\forall c$.\
		\vspace{0pt}
		\State \textbf{for} global iteration $t=1,...,T$ \textbf{do}
		\vspace{0pt}
		\State \hspace{10pt}Each client updates its model by $\mathbf{w}^{t}$. 
		\vspace{0pt}
		\State \hspace{10pt}\textbf{for} intra-set iteration $i=1,...,\tau$ \textbf{do}
		\vspace{0pt}
		\State \hspace{20pt}Each client $k \in \mathcal{V}^c$ updates its local model $\mathbf{w}_{kc}^{l+1,i,t}$ for $l = \left\{0,\ldots,L-1\right\}$ according to \eqref{my2_local}.
		\vspace{0pt}
		\State \hspace{20pt}Each client $k \in \mathcal{V}^c$ updates ${\boldsymbol \pi}^{i,t}_{kc}$ according to \eqref{my2_multiple}.
		\State \hspace{20pt}Each edge server $c$ obtains its intra-set model $\mathbf{w}^{i+1,t}_c$ according to \eqref{my2_edge}.
		\State \hspace{20pt}Each client $k \in \mathcal{V}^c$ updates as $\mathbf{w}_{kc}^{0,i+1, t} = \mathbf{w}^{i+1,t}_c$.
		\vspace{0pt}
		\State \hspace{8pt} \textbf{end}
		\State \hspace{10pt}Each edge server $c$ updates as $\mathbf{w}^{t}_c = \mathbf{w}_c^{\tau,t}$.
		\State \hspace{10pt}Each edge server $c$ and its intra-set clients update ${\boldsymbol \pi}^{t}_c$ according to \eqref{my_multiple}.
		
		\State \hspace{10pt}Cloud server obtains global model $\mathbf{w}^{t+1}$ according to \eqref{my_cloud}.
		\State \hspace{0pt} \textbf{end}
	\end{algorithmic}
	
\end{algorithm}

\vspace{-10pt}
\section{Convergence Analysis}
In this section, we analyze the convergence of {\fontfamily{lmtt}\selectfont {HierFADMM}} and {\fontfamily{lmtt}\selectfont {HierF2ADMM}} under the assumption that the number of intra-set iterations $\tau$ grows over time. We leave the analysis of the case with fixed $\tau$ for future work.

{\fontfamily{lmtt}\selectfont {HierFADMM}} and {\fontfamily{lmtt}\selectfont {HierF2ADMM}} by design can be interpreted as ADMM with an additive error, denoted by $\mathbf{e}^t_c$, arising from the inexact solution of the second update in~\eqref{my_admm}, that is
\begin{align}
\label{my_admm_inexact}
\left\{
\begin{aligned}
&\mathbf{w}^{t+1} = \arg \min_{\mathbf{w}} {\cal L}\left(\mathbf{w}, \mathbf{W}^t, \boldsymbol \Pi^t \right), \\
&\mathbf{w}^{t+1}_c = \arg \min_{\mathbf{w}_c} {\cal L}_c\left(\mathbf{w}^{t+1}, \mathbf{w}_c, \boldsymbol \pi_c^t \right) + \mathbf{e}^t_c, \forall c,\\
&{\boldsymbol \pi}^{t+1}_c = {\boldsymbol \pi}^{t}_c + \sigma_c \left(\mathbf{w}^{t+1}_c - \mathbf{w}^{t+1}\right), \forall c.
\end{aligned}
\right.
\end{align}
The additive error is different depending on whether the local training employs gradient descent ({\fontfamily{lmtt}\selectfont {HierFADMM}}) or ADMM ({\fontfamily{lmtt}\selectfont {HierF2ADMM}}).
Both algorithms thus coincides with \cite[Algorithm~1]{eckstein}, and the following result holds.

\begin{proposition}
Assume $f_{kc}$ in \eqref{localloss} to be closed, proper and convex for all $k, c$. Consider a version of {\fontfamily{lmtt}\selectfont {HierFADMM}} or {\fontfamily{lmtt}\selectfont {HierF2ADMM}} in which the number of intra-set iterations ($\tau^t$) changes over time according to $\lim_{t \to \infty} \tau^t = \infty$. Then $\mathbf{w}^t$ and $\mathbf{w}_c^t$, $\forall c$, converge to a solution of~\eqref{re_opt}.
\end{proposition}
\textit{Proof.} As discussed above, we can interpret {\fontfamily{lmtt}\selectfont {HierFADMM}} and {\fontfamily{lmtt}\selectfont {HierF2ADMM}} as the inexact ADMM in~\eqref{my_admm_inexact}. Additionally, if the number of intra-set iterations grows according to $\lim_{t \to \infty} \tau^t = \infty$, then $\lim_{t \to \infty} \mathbf{e}^t_c = \mathbf{0}$, as the intra-set updates become increasingly accurate. Therefore, we can apply \cite[Proposition~1]{eckstein} to guarantee convergence. \hfill$\square$

\vspace{-5pt}
\section{Experimental Results}
We consider the logistic regression learning task, following \cite{fedadmm, shashi}, with the parameter values provided in Table I. In this setting, each client $k \in {\cal V}^c$ has its objective function as $f_{kc}(\mathbf{w}) =$
\begin{align}
\hspace{-10pt}\frac{1}{D_{kc}} \sum_{j=1}^{D_{kc}} \left(\text{ln}\left(1+e^{\langle\mathbf{a}_{kc}^j,\mathbf{w}\rangle}\right) - b_{kc}^j\langle\mathbf{a}_{kc}^j,\mathbf{w}\rangle\right) + \frac{\lambda}{2} \|\mathbf{w}\|^2,
\end{align}
where $(\mathbf{a}_{kc}^j, b_{kc}^j \in \left\{0,1\right\})$ is the $j$-th sample in ${\cal D}_{kc}$, and $\lambda = 0.001$ is a penalty parameter. We use the Adult Census Income dataset from the UCI Machine Learning Repository to generate the samples. In the following, we compare our algorithms, {\fontfamily{lmtt}\selectfont {HierFADMM}} and {\fontfamily{lmtt}\selectfont {HierF2ADMM}}, with the conventional algorithm, {\fontfamily{lmtt}\selectfont {HierFed}}, under various scenarios.

Figs 2 and 3 show the objective for two different values, $L=1$ and $L=4$, under an i.i.d. dataset distribution across clients. It is observed that in both cases, {\fontfamily{lmtt}\selectfont {HierFADMM}} significantly outperforms {\fontfamily{lmtt}\selectfont {HierFed}}. Additionally, while {\fontfamily{lmtt}\selectfont {HierF2ADMM}} outperforms {\fontfamily{lmtt}\selectfont {HierFADMM}} with a higher number of local steps ($L=4$), its performance is not convergent when the number of local steps is minimized to 1. This is because {\fontfamily{lmtt}\selectfont {HierF2ADMM}} approximately solves the distributed minimization \eqref{step2} via ADMM, requiring a minimum number of local iterations to achieve a sufficiently accurate approximation.

In Fig. 4, the objective is shown for two scenarios with different numbers of sets, specifically $C = 3$ and $C = 6$, while the total number of clients in the system is fixed at $150$. Thus, for each case, the number of clients per cluster is given by $N_c = \frac{150}{C}, \forall c$, with $L = 4$. As illustrated, the performance gap between {\fontfamily{lmtt}\selectfont {HierFADMM}} and {\fontfamily{lmtt}\selectfont {HierF2ADMM}} widens as $C$ decreases. This occurs because more clients per cluster improve the approximation needed to solve \eqref{step2} using ADMM.

Fig. 5 displays the objective for $L=4$ under a non-i.i.d. dataset distribution across clients. In this scenario, each client holds samples from only one class, with the number of samples varying across clients. As observed, {\fontfamily{lmtt}\selectfont {HierF2ADMM}} outperforms {\fontfamily{lmtt}\selectfont {HierFADMM}}, and {\fontfamily{lmtt}\selectfont {HierFADMM}} significantly outperforms {\fontfamily{lmtt}\selectfont {HierFed}}.

\begin{table}
	\caption {Algorithm Parameters} 
	\vspace{-8pt}
	\begin{center}
		\resizebox{5.6cm}{!} {
			\begin{tabular}{| l | l | l | l | l | l | l | l | l | l | l | l |}
				
				\hline
				\hline
				{$C$}&{$N_c, \forall c$} &$\mu$& $\sigma_c, \forall c$ &$\sigma_{kc}, \forall k, c$& $\tau$ \\ \hline
				$5$& $30$ &$0.01$& $0.1$& $0.1$ & $6$ \\ \hline	
				\hline
		\end{tabular}}
	\end{center}
	\vspace{-8pt}
\end{table}
\begin{figure}[tb!]
	\vspace{0pt}
	\centering
	\includegraphics[width =1.8in]{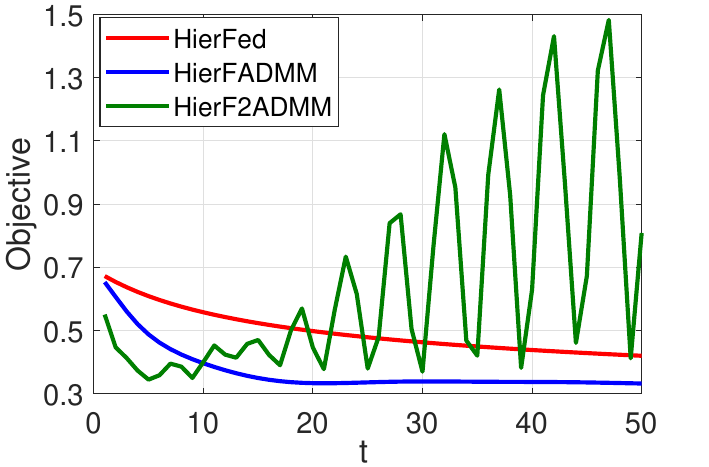} 
	\vspace{-3pt}
	\caption{Objective as a function of global iterations ($L=1$, i.i.d.)}
	\vspace{-5pt}
\end{figure}

\begin{figure}[tb!]
	\vspace{0pt}
	\centering
	\includegraphics[width =1.8in]{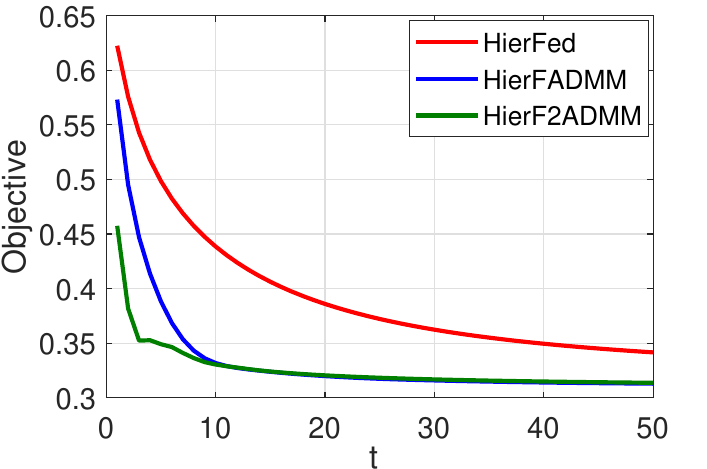} 
	\vspace{-3pt}
	\caption{Objective as a function of global iterations ($L=4$, i.i.d.)}
	\vspace{-5pt}
\end{figure}

\begin{figure}[tb!]
	\vspace{0pt}
	\centering
	\includegraphics[width =1.8in]{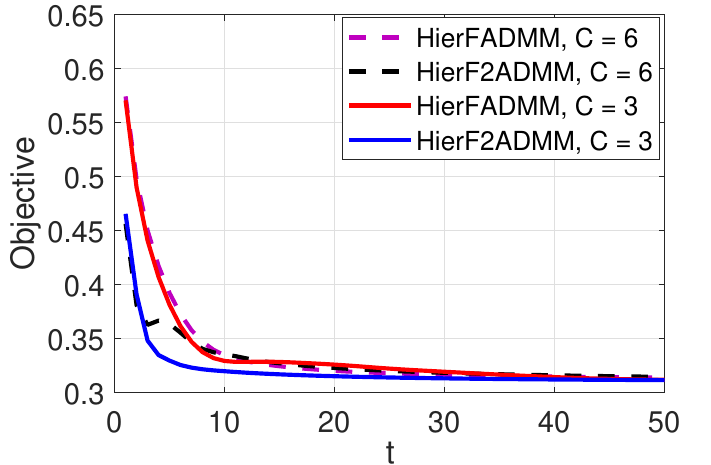} 
	\vspace{-3pt}
	\caption{Objective as a function of global iterations ($L=4$, i.i.d.)}
	\vspace{-5pt}
\end{figure}

\begin{figure}[tb!]
	\vspace{0pt}
	\centering
	\includegraphics[width =1.8in]{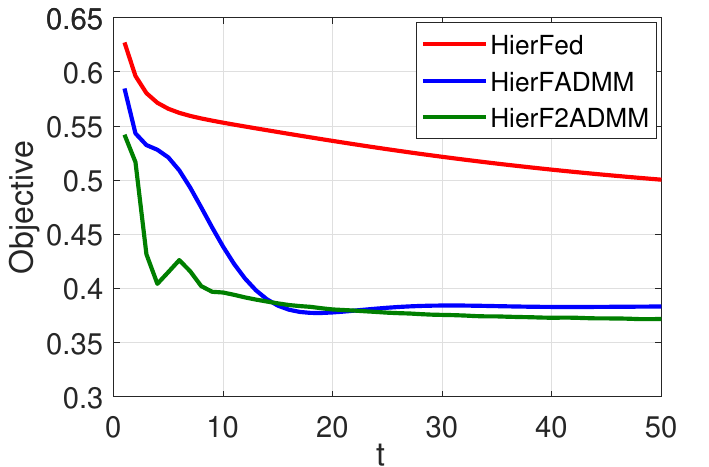} 
	\vspace{-3pt}
	\caption{Objective as a function of global iterations ($L=4$, non-i.i.d.)}
	\vspace{-15pt}
\end{figure}

\section{Conclusions}
We developed hierarchical FL based on an ADMM approach at the top layer of the hierarchy. Within this framework, we introduced two novel algorithms: hierarchical federated ADMM, which employs conventional gradient descent-based FL, and hierarchical federated double ADMM, which incorporates ADMM, at the lower layer. The new edge and cloud aggregations, coupled with local model and Lagrangian multiplier updates, in both algorithms results in enhanced learning performance and higher privacy compared to conventional hierarchical FL that relies solely on gradient descent. Our approach facilitates the integration of various other optimization methods at both lower and top layers in future implementations. Each optimization method brings its unique characteristics, enabling the combination of diverse optimization features within a single FL framework.

\appendices


\vspace{-10pt}
\begin{thebibliography}{1}
\bibitem{mcmahan}
B. McMahan, E. Moore, D. Ramage, S. Hampson, and B. A. Arcas, "Communication-efficient learning of deep networks from decentralized data," \emph{AISTATS}, pp. 1273-1282, 2017.
\bibitem{cast}
T. Castiglia, A. Das, and S. Patterson, "Multi-level local SGD: Distributed	SGD for heterogeneous hierarchical networks," \emph{ICLR}, pp. 1-36, 2021.
\bibitem{wen}
W. Wen, Z. Chen, H. H. Yang, W. Xia, and T. Q. S. Quek, "Joint scheduling
and resource allocation for hierarchical federated edge learning,"
\emph{IEEE Trans. Wireless Commun.}, vol. 21, no. 8, pp. 5857-5872, Aug. 2022.
\bibitem{gupta}
S. Gupta, W. Zhang, and F. Wang, "Model accuracy and runtime tradeoff in distributed deep learning: A systematic study," \emph{IEEE ICDM}, Barcelona, Spain, Dec. 2016.
\bibitem{lin}
F. P. C. Lin, S. Hosseinalipour, N. Michelusi, and C. G. Brinton, "Delay-aware hierarchical federated learning," \emph{IEEE Trans. Cogn. Commun. Netw.}, vol. 10, no. 2, pp. 674-688, Apr. 2024.
\bibitem{wu}
Q. Wu, X. Chen, T. Ouyang, Z. Zhou, X. Zhang, S. Yang, and J. Zhang, "HiFlash: Communication-efficient hierarchical federated learning with adaptive staleness control and heterogeneity-aware client-edge association," \emph{IEEE Trans. Parallel Distrib. Syst.}, vol. 34, no. 5, pp. 1560-1579, May 2023.
\bibitem{zhou}
X. Zhou, X. Ye, K. I. Wang, W. Liang,
N. K. C. Nair, S. Shimizu, Z. Yan, and Q. Jin, "Hierarchical federated learning with social context clustering-based participant selection for internet of medical things applications," \emph{IEEE Trans. Comput. Soc.}, vol. 10, no. 4, pp. 1742-1751, Aug. 2023.
\bibitem{zhang}
Z. Zhang, Z. Gao, Y. Guo, and Y. Gong, "Scalable and low-latency federated learning with cooperative mobile edge networking," \emph{IEEE Trans. Mobile Comp.}, vol. 23, no. 1, pp. 812-822, Jan. 2024.
\bibitem{letaief}
L. Liu, J. Zhang, S. H. Song, and K. B. Letaief, "Hierarchical federated learning with quantization: Convergence analysis and system design," \emph{IEEE Trans. Wireless Commun.}, vol. 22, no. 1, pp. 2-18, Jan 2023.
\bibitem{ourotherjournal}
S. M. Azimi-Abarghouyi and V. Fodor, "Quantized hierarchical federated learning: A robust approach to
statistical heterogeneity," under review for a journal publication, available on arXiv: https://arxiv.org/abs/2403.01540 
\bibitem{ourjournal}
S. M. Azimi-Abarghouyi and V. Fodor, "Scalable hierarchical over-the-air federated learning," \emph{IEEE Trans. Wireless Commun.}, vol. 23, no. 8, pp. 8480-8496, Aug. 2024.
\bibitem{ourconf}
S. M. Azimi-Abarghouyi and V. Fodor, "Hierarchical over-the-air federated learning with awareness of interference and data Heterogeneity," \emph{IEEE WCNC}, Dubai, UAE, April 2024.
\bibitem{prun1}
M. F. Pervej, R. Jin, and H. Dai, "Hierarchical federated learning in wireless networks: Pruning tackles bandwidth scarcity and system heterogeneity," \emph{IEEE Trans. Wirless Commun.}, vol. 23, no. 9, pp. 11417-11432, Sep. 2024.
\bibitem{bennis}
S. Liu, G. Yu, X. Chen, and M. Bennis, "Joint user association and resource allocation for wireless hierarchical federated learning with IID and non-IID data," \emph{IEEE Trans. Wireless Commun.}, vol. 21, no. 10, pp. 7852-7866, Oct. 2022.
\bibitem{tony}
S. Luo, X. Chen, Q. Wu, Z. Zhou, and S. Yu, "HFEL: Joint edge association and resource allocation for cost-efficient hierarchical federated edge learning," \emph{IEEE Trans. Wireless Commun.}, vol. 19, no. 10, pp. 6535-6548, Oct. 2020.
\bibitem{energy_resource}
R. Hamdi, A. B. Said, E. Baccour, A. Erbad,
A. Mohamed, M. Hamdi,
and M. Guizan, "Optimal resource management for hierarchical
federated learning over HetNets with
wireless energy transfer," \emph{IEEE Internet Things J.}, vol. 10, no. 19, pp. 15299-15309, Oct. 2023.
\bibitem{feduc}
Q. Ma, Y. Xu, H. Xu, J. Liu, and L. Huang, "FedUC: A unified clustering approach for
hierarchical federated learning," \emph{IEEE Trans. Mob. Comput.}, early access.
\bibitem{boyd}
S. Boyd and L. Vandenberghe, \emph{Convex Optimization.} Cambridge Univ. Press, 2004.
\bibitem{fedadmm}
S. Zhou and G. Y. Li, "Federated learning via inexact ADMM," \emph{IEEE Trans. Pattern Anal. Mach. Intell.}, vol. 45, no. 8, pp. 9699-9708, Aug. 2023.
\bibitem{shashi}
S. Kant, J. M. B. da Silva, G. Fodor, B. Goransson, M. Bengtsson, and C. Fischione, "Federated learning using three-operator ADMM," \emph{IEEE J. Sel. Top. Signal Process.}, vol. 17, no. 1, pp. 205-221, Jan. 2023.
\bibitem{eckstein}
J. Eckstein and W. Yao, "Relative-error approximate versions of Douglas-Rachford splitting and special cases of the ADMM," \emph{Math. Program.}, vol. 170, no. 2, pp. 417-444, Aug. 2018.
\end{thebibliography}
\end{document}